\documentclass[10pt,journal,compsoc]{IEEEtran}
\ifCLASSOPTIONcompsoc
  \usepackage[nocompress]{cite}
\else
  \usepackage{cite}
\fi
\ifCLASSINFOpdf
\else
\fi
\usepackage{hyperref}       
\hypersetup{colorlinks=true}
\usepackage{url}            
\usepackage{booktabs}       
\usepackage{amsfonts}       
\usepackage{nicefrac}       
\usepackage{microtype}      
\usepackage{xcolor}         
\usepackage{amsmath}
\usepackage{multirow}
\usepackage{float}
\usepackage{color, colortbl}
\usepackage{xcolor}
\usepackage{rotating}
\usepackage{graphicx,wrapfig}
\usepackage[export]{adjustbox}
\usepackage{enumitem}
\usepackage[noend]{algorithm2e}
\usepackage{pifont}

\definecolor{dgreen}{RGB}{1,150,74}
\definecolor{coralpink}{rgb}{0.97, 0.51, 0.47}

\newcommand\red[1]{\textcolor{red}{#1}}
\newcommand\dgreen[1]{\textcolor{dgreen}{#1}}
\newcommand\gray[1]{\textcolor{gray}{#1}}
\newcommand\cyan[1]{\textcolor{cyan}{#1}}

\newcommand\down[1]{\textcolor{dgreen}{$\downarrow{#1}$}}

\newcommand\ul[1]{\underline{#1}}

\newcommand{\cmark}{\ding{51}}%
\newcommand{\xmark}{\ding{55}}%
\begin{document}
%
\title{FSD V2: Improving Fully Sparse 3D Object Detection with Virtual Voxels}
%
%
%
%

\author{Lue Fan,
        Feng Wang,
        Naiyan Wang,
        and Zhaoxiang Zhang
\IEEEcompsocitemizethanks{
\IEEEcompsocthanksitem{Lue~Fan and Zhaoxiang~Zhang are with Center for Research on Intelligent Perception and Computing (CRIPAC), National Laboratory of Pattern Recognition (NLPR), Institute of Automation, Chinese Academy of Sciences (CASIA), Beijing 100190, China. E-mail: \{fanlue2019, zhaoxiang.zhang\}@ia.ac.cn.}
\IEEEcompsocthanksitem{
          Feng~Wang and Naiyan~Wang are with TuSimple, Beijing 100020,
          China. E-mail: \{feng.wff, winsty\}@gmail.com.}
}
}

%
%

\markboth{Journal of \LaTeX\ Class Files,~Vol.~14, No.~8, August~2015}%
{Shell \MakeLowercase{\textit{et al.}}: Bare Demo of IEEEtran.cls for Computer Society Journals}
%



\IEEEtitleabstractindextext{%
\begin{abstract}
LiDAR-based fully sparse architecture has garnered increasing attention. FSDv1 stands out as a representative work, achieving impressive efficacy and efficiency, albeit with intricate structures and handcrafted designs.
In this paper, we present FSDv2, an evolution that aims to simplify the previous FSDv1 while eliminating the inductive bias introduced by its handcrafted instance-level representation, thus promoting better general applicability.
To this end, we introduce the concept of \textbf{virtual voxels}, which takes over the clustering-based instance segmentation in FSDv1.
Virtual voxels not only address the notorious issue of the Center Feature Missing problem in fully sparse detectors but also endow the framework with a more elegant and streamlined approach.
Consequently, we develop a suite of components to complement the virtual voxel concept, including a virtual voxel encoder, a virtual voxel mixer, and a virtual voxel assignment strategy. Through empirical validation, we demonstrate that the virtual voxel mechanism is functionally similar to the handcrafted clustering in FSDv1 while being more general.
We conduct experiments on three large-scale datasets:
\textbf{Waymo Open Dataset}, \textbf{Argoverse 2} dataset and \textbf{nuScenes} dataset.
Our results showcase state-of-the-art performance on all three datasets, highlighting the superiority of FSDv2 in long-range scenarios and its general applicability to achieve competitive performance across diverse scenarios.
Moreover, we provide comprehensive experimental analysis to elucidate the workings of FSDv2. To foster reproducibility and further research, we have open-sourced FSDv2 at \url{https://github.com/tusen-ai/SST}.
\end{abstract}

}

\maketitle

\IEEEdisplaynontitleabstractindextext

%
\IEEEpeerreviewmaketitle

\IEEEraisesectionheading{\section{Introduction}\label{sec:introduction}}

\IEEEPARstart{F}ully sparse detectors~\cite{fsd, fsd++, voxelnext, flatformer, iassd} have showcased remarkable efficiency and efficacy in LiDAR-based 3D object detection, particularly excelling in long-range scenarios. 
This superiority holds significant promise for enhancing the safety of autonomous driving.
\par
Nonetheless, the development of effective fully sparse detectors faces a significant obstacle in the form of the \textbf{Center Feature Missing} (CFM)~\cite{fsd, votenet}, which stems from the sparse nature of point clouds.
CFM denotes that central features of objects are frequently absent in fully sparse architectures since the point clouds predominantly distribute on the surfaces of objects. 
To circumvent this challenge, the pioneering work FSD\cite{fsd} employs an instance-level representation: using clustering to obtain instances and then extract instance-level features for bounding-box prediction, thereby obviating the need for object center features.
Although effective, the instance-level representation introduces strong inductive bias, impeding the general applicability.
For example, the clustering part in instance segmentation necessitates pre-defined and handcrafted distance thresholds for each category, and it is non-trivial to find optimal values.
Moreover, such a rule-based clustering is prone to erroneously recognize multiple objects as a single entity in crowded scenes, leading to false negatives.
Although turning to meticulously designed instance segmentation methods may alleviate this issue, it is not an essential solution and increases the complexity.

\begin{figure}[t]
	\centering
	\includegraphics[width=0.99\linewidth]{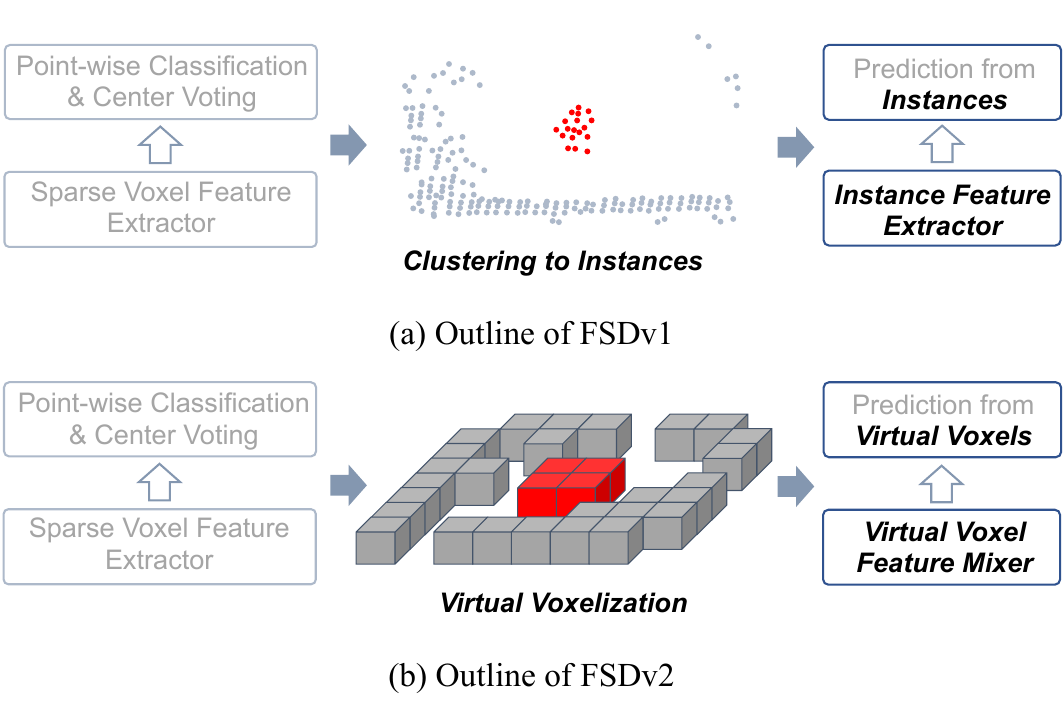}
	\caption{
	{Conceptual comparison with FSDv1 and the proposed FSDv2. FSDv1 relies on handcrafted clustering-based instance-level representation which introduces inductive bias, reducing the general applicability. Instead, FSDv2 replaces clusters with virtual voxels (\red{red} voxels) from the voted centers (\red{red} points).}
 }
	\label{fig:fig1}
\end{figure}

In this paper, we propose FSDv2, an approach designed to streamline the structure of FSDv1 by discarding its instance-level representation, pursuing better general applicability.
As depicted in Fig. \ref{fig:fig1}, compared with FSDv1, the primary advancements in FSDv2 are concentrated in the structures after center voting.
In FSDv2, we introduce the concept of \textbf{virtual voxels}, which serves as a replacement for the instances in FSDv1. These virtual voxels are derived by voxelizing the voted centers.
We use the term ``virtual'' because the voted centers are artificial and not the real points obtained by sensors.
Then the virtual voxels are passed into a light-weight sparse \textbf{Virtual Voxel Mixer} (VVM) for feature enhancement.
The VVM is necessary because a virtual voxel near an object center may only contain a partial set of the voted centers when the center voting is not that good.
So we need VVM to aggregate the features of different virtual voxels belonging to a specific object, resulting in better features covering the whole instance.
By this design, VVM intuitively mimics the behavior of instance-level feature extraction (i.e., SIR in Sec.~\ref{sec:preliminary}) in FSDv1 but does not explicitly generate instances, avoiding introducing strong handcrafted inductive bias.
After VVM, the virtual voxel features are enhanced and we predict the bounding box directly from the virtual voxels.

The virtual voxel mechanism, despite its simplicity, assumes a versatile role within FSDv2.
(1) By replacing the clustering-based instance segmentation in FSDv1, virtual voxelization significantly simplifies the architecture and reduces inductive bias, thereby enhancing the methodology's general applicability.
(2) Virtual voxels serve as ``anchor points'' to predict the bounding boxes.
Since the virtual voxels are around the object centers, using them as anchor points greatly reduces the variance of regression targets.
(3) If we have perfect point classification and center voting, the number of virtual voxels is the same as the number of objects.
Thus, prediction from virtual voxels greatly mitigate the imbalance between positive and negative samples and between objects containing a highly variable number of points.
\par
We summarize our contributions as follows.
\begin{itemize}
    \item We propose FSDv2, a fully sparse LiDAR-based 3D object detector, specifically designed to improve the performance of its FSDv1 predecessor.
    FSDv2 replaces the instance-level representation in FSDv1 with virtual voxels.
    This design overhaul streamlines the pipeline and removes the inductive bias arising from clustering-based instance segmentation.
    \item 
    We conduct experiments on three large-scale benchmarks: Waymo Open Dataset, Argoverse 2 dataset, and nuScenes dataset.
    State-of-the-art performance is reported on these datasets, revealing the general applicability of FSDv2. 
    \item
    The simplicity of FSDv2's architecture and its general applicability position it as a strong and accessible baseline model.
    By open-sourcing FSDv2, we aim to encourage and facilitate further exploration and advancements in the domain of fully sparse detectors.
\end{itemize}
\section{Related Work}
\subsection{Dense Detectors}
The pioneering work of VoxelNet~\cite{voxelnet} introduced the use of dense convolutions for voxel feature extraction, which yielded competitive performance in 3D object detection.
However, applying dense convolutions directly to 3D voxel representations posed efficiency challenges due to its computational complexity.
To address this limitation, subsequent approaches like PIXOR~\cite{pixor} and PointPillars~\cite{pointpillar} takes a different approach by adopting 2D dense convolutions on Bird's Eye View (BEV) feature maps, resulting in a significant improvement in computational efficiency.
We refer to these detectors as \textbf{dense detectors}, as they convert the sparse point cloud into dense feature maps.

\subsection{Semi-dense Detectors}
Different from dense detectors, semi-dense detectors incorporate both sparse features and dense features.
SECOND~\cite{second} is the pioneering work to adopt sparse convolution to extract the 3D sparse voxel features and then convert sparse features to dense BEV feature maps for the integration with 2D detection head~\cite{ssd, fasterrcnn, centernet}.
The prominence of semi-dense detectors has surged further with the introduction of CenterPoint~\cite{centerpoint}, which established a strong baseline.
Based on this paradigm, many studies~\cite{transfusion, afdetv2, centerformer} improve the performance from different perspectives.
Another hot trend aims to enhance the sparse backbone by transformer architecture~\cite{transformer}, such as~\cite{votr, sst, voxset, flatformer, dsvt, swformer, mssvt}.
Moreover, there is a line of work attaches a second stage for fine-grained feature extraction and proposal refinement~\cite{parta2, pvrcnn, pvrcnnpp, voxelrcnn, pyramidrcnn}.
Despite leveraging the sparse backbone, these methods are hard to scale up to long-range scenarios due to the requirement of dense feature maps in the detection head.

\subsection{Fully Sparse Detectors}
Some early work is purely based on points, which are born to be fully sparse due to voxelization is not included.
PointRCNN~\cite{pointrcnn} stands out as the pioneering work in developing a purely point-based detector.
Building upon this foundation, 3DSSD~\cite{3dssd} accelerated the point-based approach by eliminating the feature propagation layer and refinement module. 
VoteNet~\cite{votenet} first makes a center voting and then generates proposals from the voted center achieving better precision.
As a point-based method, IA-SSD~\cite{iassd} is concentrated more on the foreground, attaining remarkable efficiency. 
Albeit many efforts to accelerate the point-based method, the time-consuming neighborhood query is still unaffordable in large-scale point clouds (more than 100k points per scene).
FSDv1~\cite{fsd} is the first work positioning the concept of fully sparse architecture at the forefront.
It develops a fully sparse pipeline, sidestepping the issue of the center feature missing and avoiding time-consuming operations in purely point-based methods.
FSDv1 is adapted to process tracklet data with large spatial spans for offline auto-labeling in CTRL~\cite{ctrl}.
VoxelNeXt~\cite{voxelnext} further simplifies the fully sparse architecture with purely voxel-based designs and achieves competitive performance on detection and tracking.
FlatFormer~\cite{flatformer} combines a novel transformer-based sparse backbone with the detection head in FSDv1, resulting in competitive performance and impressive efficiency.
Based on FSDv1, FSD++~\cite{fsd++} improves the sparsity by removing temporal redundancy.
This line of progress highlights the potential of fully sparse architecture in revolutionizing 3D object detection methodologies.

\subsection{Treatments to Center Feature Missing}
The issue of the Center Feature Missing has its origins in VoteNet~\cite{votenet}.
VoteNet points out that directly predicting bounding box parameters from surface points is challenging.
So it makes center voting and then makes predictions based on the voted key points.
Such a solution is sub-optimal since it needs clustering operations such as ball query.
FSDv1~\cite{fsd} builds upon the basic idea of clustering but proposes a more effective and efficient cluster-level feature extraction module.
Despite achieving competitive performance, both VoteNet and FSDv1 encounter inductive bias due to the clustering operation, which hinders generalization.
Other treatments to center feature missing are proposed in RSN~\cite{rsn}, SWFormer~\cite{swformer}, and VoxelNeXt~\cite{voxelnext}.
RSN and VoxelNeXt assign positive labels to the nearest voxels of object centers, mitigating the problem to some extent, but the essential issue of centers being empty remains.
SWFormer proposes voxel diffusion to expand the non-empty regions, which involves a max-pooling operation and conceptually resembles the dilation operation in morphology.
Although voxel diffusion is proven effective, it greatly reduces spatial sparsity, potentially impairing the efficiency of models.

\begin{figure*}[t]
	\centering
	\includegraphics[width=0.99\linewidth]{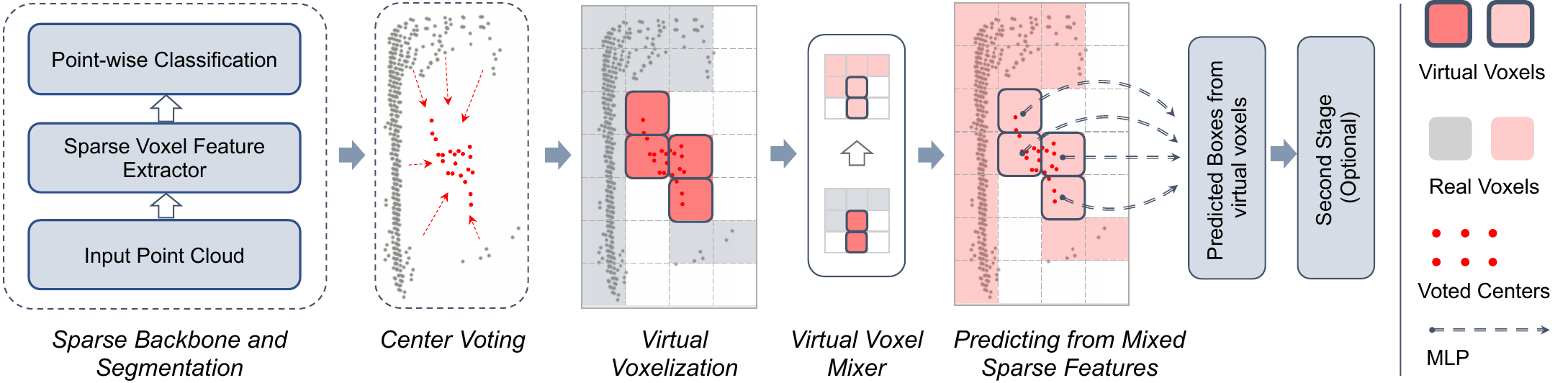}
	\caption{
	\textbf{Overview of FSDv2.} The components before virtual voxelization is the same as FSDv1. Afterward, we employ virtual voxelization on both the original real points and the voted centers.
    Then we mix the features of virtual voxels (the four \textbf{outlined} \red{red} voxels) and real voxels (\gray{gray}) via a Virtual Voxel Mixer for enhancement.
    We use \textcolor{coralpink}{light red} to show the features after mixing.
    Eventually, predictions are made from the virtual voxels.
 }
	\label{fig:pipeline}
\end{figure*}
\section{Preliminary}
\label{sec:preliminary}
For a clear understanding of the evolution from FSDv1 to FSDv2, we briefly introduce the structure of FSDv1 in this section.
\subsection{Overall Design of FSDv1}
 As Fig. \ref{fig:fig1} shows, FSDv1 can be roughly divided into three parts.
(1) \textbf{Point Feature Extraction}. Sparse voxel feature extractor~\cite{sst, parta2} serves as the backbone for voxel feature extraction.
Subsequently, an MLP-based neck converts voxel features into point features.
Finally, a lightweight point-wise MLP is employed to perform point-wise classification and center voting.
(2) \textbf{Clustering}. In this step, Connected Component Labeling (CCL) is applied to the voted centers to cluster points into instances. 
(3) \textbf{Instance feature extraction and box prediction}. FSDv1 then uses sparse operators to extract instance features for bounding box prediction. We present the details of instance feature extraction in the following.

\subsection{Sparse Instance Recognition}
\label{sec:sir}
The core operator in FSDv1 for efficient instance-level extraction is Sparse Instance Recognition (SIR).
In FSDv2, this part will be removed and further adapted to a virtual voxel encoder (Sec.~\ref{sec:virtual_voxel_encoder}), so we briefly introduce its structure here.
\par
In particular, initial instance point features are first passed into an MLP.
Then the output point features are aggregated into instance-wise features by instance-wise max-pooling.
The obtained instance features are then broadcasted to each point within the instance and concatenated with the pre-aggregated point features.
The concatenated point features are further passed through another MLP to reduce channel dimension.
The process above could be applied iteratively.
The eventual instance features are obtained by max pooling the point features without broadcasting back to points again.
Finally, box predictions are made from the eventual instance features.
\par
In a nutshell, SIR is similar to a couple of PointNet~\cite{pointnet} layers. Given a point set, the SIR module efficiently extracts the set feature. In FSDv1, it is adopted to extract instance features, and we use it to extract voxel features in Sec.~\ref{sec:virtual_voxel_encoder}.

\subsection{Limitations}
The core idea behind FSDv1 is clustering the points to instances and making predictions from the instance features instead of empty object centers.
Although the clustering sidesteps the issue of the center feature missing, its handcrafted designs introduce extra inductive bias and impair generalization capabilities.
Therefore, in this paper, we propose the use of virtual voxels as an alternative to clustering.

\section{Methodology}
\subsection{Overall Architecture}
Fig.~\ref{fig:pipeline} outlines the pipeline of FSDv2. Initially, a sparse voxel feature extractor serves as the backbone and then MLP-based module is employed for point-wise classification and point-wise center voting.
So far, the FSDv2 shares these structures with FSDv1, while things become different after the center voting.
FSDv2 then applies virtual voxelization instead of clustering, presented in Sec.~\ref{sec:virtual_voxels}.
Subsequently, we propose a virtual voxel mixer to mix features of different virtual voxels, presented in Sec.~\ref{sec:mixer}.
The synergy of virtual voxelization and virtual voxel mixer contributes a similar effect to SIR in previous FSDv1, and an intuitive discussion on this is provided in Sec.~\ref{sec:discussion}.
The virtual voxels are adopted for bounding box prediction, so we elaborate on the label assignment of virtual voxels in Sec.~\ref{sec:assignment}.
The design of the detection head is presented in Sec.~\ref{sec:head}.

\subsection{Virtual Voxelization}
\label{sec:virtual_voxels}
\subsubsection{Virtual Voxels}
Following the previous FSDv1, the sparse voxel encoder (i.e., backbone) extracts the voxel features and maps the voxel features to point features.
Then we conduct point-wise classification and center voting based on the point features.
In what follows, instead of clustering points into instances in FSDv1, FSDv2 creates virtual voxels using the voted centers.
In particular, for each foreground point, we decode its voted object center from the predicted offset, obtaining voted centers.
Then we apply voxelization to the union of voted centers and original real points.
In this context, \textbf{Virtual voxels} refer to those voxels containing \textbf{at least one} voted center, while voxels containing only original real points are referred to as \textbf{real voxels}.
\par
Although the number of voted centers could be considerably large, there are usually only a few virtual voxels for two reasons.
(1) The voted centers are often tightly closed to each other. Thus, if the network makes perfect center voting, the number of virtual voxels equals the number of objects. 
(2) The size of virtual voxels is relatively larger than commonly used voxel sizes in the early stage since the backbone has captured the fine-grained information.
In a nutshell, the spatial sparsity degradation caused by virtual voxels is negligible.
\par
After this step, we obtain virtual and real voxels containing raw point coordinates and attributes (e.g., intensity). In the following, we proceed to encode the feature of each voxel.

\subsubsection{Virtual Voxel Feature Encoding}
\label{sec:virtual_voxel_encoder}
We introduce a \textbf{virtual voxel encoder (VVE)} to encode the virtual voxel features.
In essence, we inherit the SIR structure (Sec.~\ref{sec:sir}) in FSDv1 for virtual voxel feature encoding.
The main difference lies in that SIR treats the instance cluster as a point group and extracts the instance feature while VVE treats points in a voxel as a group to extract the voxel feature.
In particular, before voxel encoding, we need to first generate features for each voted center.
Since each vote center is predicted from a certain real point, voted centers copy the features from their corresponding real points, which are encoded by the sparse backbone.
Moreover, we append the voting offsets to voted centers as additional features to distinguish them from the real points.
For real points, we pad zeros as pseudo offsets to maintain consistent feature dimensions. 
After generating the initial point-wise features, the virtual voxel feature encoder encodes voxel features by the SIR module in Sec.~\ref{sec:sir}, which can be viewed as a couple of PointNet layers.
Please note that despite using the term "virtual," the virtual voxel encoder actually encodes features of \textbf{both} virtual voxels and real voxels.
This is because we need the real voxel features to enhance the performance in the following Sec.~\ref{sec:mixer}.

\subsection{Virtual Voxel Mixer}
\label{sec:mixer}
So far, we attain different types of features, such as virtual voxel features, real voxel features, and multi-scale features from the SparseUNet backbone.
We could mix them together by a lightweight sparse network to unify and enhance features, which we term as \textbf{virtual voxel mixer (VVM)}.
In Sec.~\ref{sec:mix1}, Sec.~\ref{sec:mix2}, and Sec.~\ref{sec:mix3}, we present the motivations and basic ideas to mix different types of features.
In Sec.~\ref{sec:mixer_structure}, we propose the specific network structure for feature mixing.
\subsubsection{Mixing Virtual Voxel Features}
\label{sec:mix1}
If the voting in an object is mediocre, there are usually multiple virtual voxels around the object center.
As illustrated in Fig.~\ref{fig:voting_cases}(b), each virtual voxel contains partial information about the object.
And these virtual voxels have not interacted with each other since the voxel encoding (Sec.~\ref{sec:virtual_voxel_encoder}) is applied within each voxel.
So it is necessary to enable feature interaction between virtual voxels in such mediocre cases.

\subsubsection{Mixing Virtual and Real Voxel Features}
\label{sec:mix2}
In addition to exclusively mixing the virtual voxel features, VVM could also mix virtual voxel features with real voxel features for further enhancement.
The motivation behind this stems from that virtual voxels are generated from predicted foreground points, which may occasionally miss certain foreground points, leading to potential information loss.
We emphasize that the previous FSDv1 has no similar mechanism to remedy the information loss caused by inaccurate point classification.

\subsubsection{Mixing Multi-scale Features}
\label{sec:mix3}
In FSDv1, the single-scale output features SparseUNet backbone serve as the driving force for all subsequent components, including SIR and final prediction.
However, we empirically observed that such single-scale output features might not be sufficient, particularly in scenarios with highly diverse categories. 
So we propose to aggregate multi-scale features via VVM.
More specifically, the multi-scale features consist of (1) multi-scale real voxel features from different layers of the SparseUNet decoder, and (2) virtual/real voxel features in Sec.~\ref{sec:mix1} and Sec.~\ref{sec:mix2}.
Unlike multi-scale aggregations in image tasks, we cannot directly employ addition or feature concatenation along channel dimension since the features are spatially sparse, exhibiting irregular positions.
Thus we present a unique solution as follows. 
\par
For clarity, we represent the sparse feature with stride $s$ as $F_{s}$, with a shape of $[N_s, C_s]$, where $N_s$ denotes the number of features (i.e., voxels), and $C_s$ represents the channel dimension. Correspondingly, the integer spatial coordinates of voxels are denoted as $I_{s}$ with a shape of $[N_s, 3]$.
$I_{s}^{\tilde{s}}$ is the coordinate converted from stride $s$ to stride $\tilde{s}$. 
We let $F_{1}$ denote the features obtained by virtual voxelization.
In this sense, $s$ is the relative stride to virtual voxel features.
For multi-scale mixing, we first convert $I_{s}$ to $I_{s}^{1}$ by:
\begin{equation}
    I_{s}^{1} = I_{s} \times s + \left\lfloor\frac{s}{2}\right\rfloor.
    \label{eq:indices_conversion}
\end{equation}
This conversion allows us to put the $F_{s}$ to proper positions in the voxelized space with stride 1, which actually is the stride of virtual voxels.
So the aggregated sparse features and voxel coordinates can be denoted as 
\begin{equation}
    F_{agg} = \left [ \texttt{Lin}(F_{1}),  \texttt{Lin}(F_{2}), \mathellipsis, \texttt{Lin}(F_{L})\right ],
    \label{eq:multiscale_features}
\end{equation}
and
\begin{equation}
    I_{agg} = \left [ I_{1},  I_{2}^{1}, \mathellipsis, I_{L}^{1}\right ],
    \label{eq:multiscale_indices}
\end{equation}
where $[\cdot]$ denotes concatenation along the dimension of voxel numbers, and $\texttt{Lin}$ denotes a linear layer to obtain consistent channel dimensions.
Intuitively, Eq.~\ref{eq:multiscale_features} and Eq.~\ref{eq:multiscale_indices} imply that we position sparse voxel features with different strides into the voxelized space with stride 1, following the conversion rule in Eq.~\ref{eq:indices_conversion}, while retaining the original feature values unchanged. 
\par
\begin{figure}[ht]
	\centering
	\includegraphics[width=0.99\linewidth]{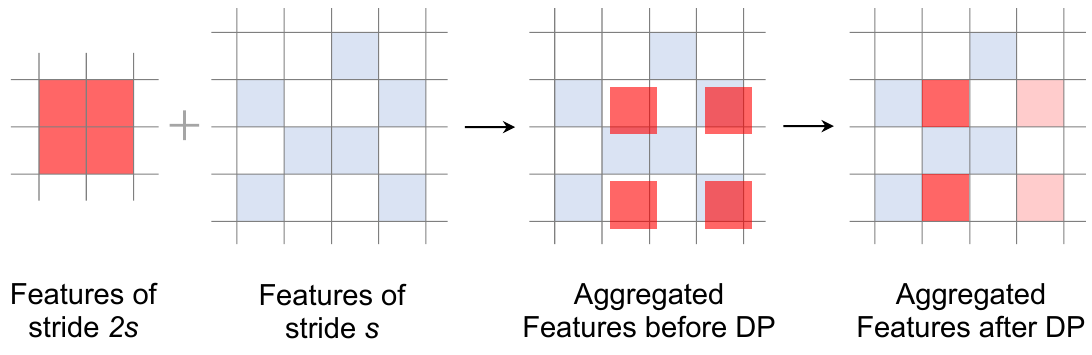}
	\caption{\textbf{Illustration of mixing multi-scale features.}
 The \textcolor{coralpink}{light red} voxel in the rightmost figure indicates they are the average-pooling results of red voxels and light blue voxels sharing the same positions. The average pooling is implemented by the dynamic pooling (DP) operator in FSDv1.
 }
	\label{fig:multiscale}
\end{figure}

An important property worth noting here is that $I_{agg}$ may contain duplicated elements since there are voxels from different scales sharing the same spatial coordinate. 
So we use the dynamic pooling operator $\texttt{DP}$ in~\cite{fsd} to unique the duplicate coordinates.
The features of duplicated voxels are averaged into a single feature.
The process is illustrated in Fig.~\ref{fig:multiscale}.
And the unique multi-scale features are together passed into the mixer for feature mixing.

\subsubsection{Model Structure of VVM}
\label{sec:mixer_structure}
We have many options to instantiate a VVM, such as various spconv-based networks~\cite{second, centerpoint, parta2, voxelnext} and the emerging sparse transformers~\cite{votr, sst, dsvt, flatformer, mssvt, voxset}.
Here we opt for a straightforward lightweight SparseUNet, which actually is the lightweight version of our backbone.
The SparseUNet is highly efficient, featuring only three stages with different strides, each consisting of two sparse convolution layers.
Due to its ``encoder-decoder'' structure, the output voxels have the same spatial distribution and resolution as the input voxels.

\subsection{Discussion: Cluster v.s. Virtual Voxel}
\label{sec:discussion}
Virtual voxelization and VVM take the place of the point clustering and SIR module in FSDv1.
To gain deeper insights into the evolution from FSDv1 to FSDv2, we qualitatively compare the behaviors of these components in different cases: when the model makes \emph{consistent} center voting and when it makes \emph{inconsistent} voting.
\par
\begin{figure}[h]
	\centering
	\includegraphics[width=1.0\linewidth]{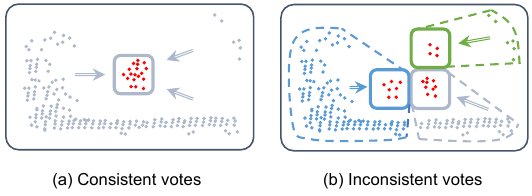}
	\caption{
	\textbf{Two kinds of voting cases.} \red{Red} points are the voted centers.
        \textbf{Left}: model makes consistent votes, so all the voted centers fall into a single virtual voxel. In this case, encoding this virtual voxel is similar to encoding the complete instance.
        \textbf{Right}: each virtual voxel corresponds to different parts of the instance.
        The virtual voxel mixer enables the interaction between different virtual voxels, which also indirectly encodes the complete instance.
 }
	\label{fig:voting_cases}
\end{figure}

\noindent \textbf{Consistent votes}
When the model makes consistent votes, all the voted centers of an object fall into a single virtual voxel.
Assuming that the clustering in FSDv1 is also nearly perfect,
under this condition, the proposed virtual voxelization is similar to the clustering-based instance representation in FSDv1 since the virtual voxel encoder and SIR share the same structure as demonstrated in Sec.~\ref{sec:virtual_voxel_encoder}.
We illustrate this in Fig.~\ref{fig:voting_cases} (a).

\noindent \textbf{Inconsistent votes}
When the model makes inconsistent votes, multiple virtual voxels surround the object center and each of them may only encode a partial shape of the object.
However, the virtual voxel mixer enables the interaction between these virtual voxels, indirectly encoding the complete shape of the object.
In this way, the synergy of virtual voxelization and mixer achieves similar functionality to SIR even though the votes are inconsistent. Fig.~\ref{fig:voting_cases} (b) illustrates our analysis.
\par
In conclusion, the proposed virtual voxel mechanism inherits the core functionality of FSDv1 and further removes the explicit handcrafted instance-level representation, making the model more simple and general.

\subsection{Virtual Voxel Assignment}
\label{sec:assignment}
So far, the virtual voxels contain well-encoded features and are positioned around the object centers.
Thus, it is a natural choice to predict bounding boxes from these virtual voxels.
Here we present how to assign labels for the virtual voxels.
\subsubsection{Potential Design Choices}
\label{sec:potential_assignments}
Since the virtual voxels usually fill the empty object centers, the popular center-based assignment~\cite{centernet, centerpoint} is a potential solution.
The conventional anchor-based assignment~\cite{second, pointpillar} is also feasible.
Recently, some methods with sparse detection heads such RSN~\cite{rsn} and VoxelNeXt~\cite{voxelnext} opt for assigning the nearest voxel of each object centers.
Although these approaches work well in certain scenarios, they are sub-optimal for virtual voxels due to the following reasons.
\begin{itemize}
    \item Virtual voxel does not always fill the object centers, especially for large and distant objects. Therefore, directly adopting the center-based assignment is not feasible.
    \item Anchor-based requires class-aware hyper-parameters like the predefined anchor sizes.
    These hyper-parameters contradict our design goal to improve general applicability and are tricky to be well-tuned in datasets with a considerable number of categories such as Argoverse 2~\cite{argo2}, which contains 30 categories with highly variable sizes.
    \item The nearest assignment may cause ambiguity and impede the optimization.
    This is because there might be multiple virtual voxels near the object centers, but only one can be selected in the nearest assignment.
    Moreover, for objects containing few points, only assigning a single virtual voxel as positive makes it likely to be a false negative.

\end{itemize}

\subsubsection{Our Solution: Voxel-in-box Assignment}
In FSDv2, we adopt a more straightforward strategy called \textbf{voxel-in-box assignment}, where \textbf{all the} virtual voxels falling within a bounding box are assigned as positive. We argue that this strategy is well-suited for FSDv2 due to the following reasons:
\begin{itemize}
    \item The number of virtual voxels of an object is much less than its real voxels. So even if we assign all the inside virtual voxels as positive, it is unlikely to cause a severe imbalance between objects containing different numbers of points. Instead, it potentially increases the recall of objects containing few points.
    \item Virtual voxels are near the object centers. So the inclusion of all virtual voxels does not lead to a large variance in regression targets. 
    \item In the 3D physical world, accurately annotated bounding boxes do not overlap with each other, and the sparsity of point clouds typically ensures that bounding boxes do not contain background clutter.
    These properties make voxel-in-box a reliable criterion for determining whether a voxel should be assigned to a certain object as positive.
    It also explains why the straightforward voxel-in-box assignment could work in 3D while image-based 2D object detection requires sophisticated strategies~\cite{fasterrcnn, fcos, atss, autoassign, detr}.
\end{itemize}

\subsubsection{Define the Position of Virtual Voxel}
To determine if a voxel is inside a box, we first need to define the position of a virtual voxel.
A straightforward strategy is to adopt the geometric center of a voxel as its position.
However, the strategy is inaccurate and even ambiguous when it comes to tiny objects.
For instance, the \textbf{bollard} and \textbf{traffic cone} classes in Argoverse 2 are only around $0.1m \times 0.1m$ in Bird Eye's View, which is significantly smaller than the default virtual voxel size of $0.5m \times 0.5m \times 0.5m$.
Thus it is hard to judge if such a large voxel is inside the tiny object.
\par
\begin{figure}[h]
	\centering
	\includegraphics[width=0.99\linewidth]{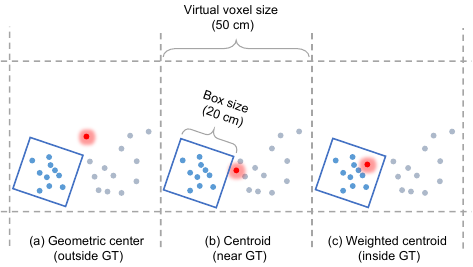}
	\caption{
	\textbf{Conceptual illustration of weighted voxel centroid for label assignment.}
 \cyan{Blue} points are foreground.
 We take tiny objects (\cyan{blue} boxes) as examples.
 (a) Although the virtual voxel fully contains the object, its geometric center is outside of the GT box, so it is assigned as negative.
 (b) The centroid of points in the virtual voxel is closer to the object, but still outside of the box (not really in practice, just for illustration).
 (c) The weighted centroid is further closer to the foreground points and inside the box, resulting in a positive label.
 }
	\label{fig:assignment}
\end{figure}

To address the issue, we take point distribution in a voxel into account. 
Specifically, the position of a virtual voxel is defined as the weighted centroid of its containing points.
Thus, \emph{a virtual voxel is inside a bounding box only if the centroid is inside the box, regardless of the virtual voxel size}. Fig.~\ref{fig:assignment} illustrates three typical strategies.
Formally, the weighted centroid is defined as:
\begin{equation}
    \bar{\mathbf{x}} = \frac{\sum_{i=0}^{N-1}\texttt{I}(\mathbf{x_i})\mathbf{x_i}}{\sum_{i=0}^{N-1}\texttt{I}(\mathbf{x_i})},
\end{equation}
where
\begin{equation}
   \texttt{I}(\mathbf{x}) = 
       \begin{cases}
        1 & \text{if } \mathbf{x} \in \mathbb{F}\\
        \alpha & \text{if } \mathbf{x} \not\in \mathbb{F}
    \end{cases}.
    \label{eq:weighted_alpha}
\end{equation}
In the equations above, $\mathbf{x_i}$ is the coordinate of the $i$-th point in a voxel, and $\mathbb{F}$ is the foreground subset of all points\footnote{Here ``all points'' refers to the union of raw points and voted centers.}, and $N$ is the number of containing points of a voxel.
$\alpha \in [0, 1]$ is a weight factor of background points.
This weighting strategy indicates that the position of a voxel is primarily determined by its containing foreground points, while also slightly influenced by background points.

\subsection{Virtual Voxel Head}
\label{sec:head}
Virtual voxel features from VVM are further passed into a couple of MLPs for final box prediction.
Following CenterPoint~\cite{centerpoint}, we partition all the categories into different groups, and the details are presented in Sec.~\ref{sec:variants}.

\par
With the assigned labels for each virtual voxel (Sec.~\ref{sec:assignment}), we design the classification and regression branches as follows.
We adopt Focal Loss~\cite{focalloss} with default hyper-parameters for classification.
We use L1Loss in the regression branch. The regression target is parameterized as $(\Delta x, \Delta y, \Delta z, \log{l}, \log{w}, \log{h}, \sin{\theta}, \cos{\theta})$.
$\Delta x$ is the $x$-offset from the bounding box gravity center to \emph{geometric} center of the assigned virtual voxel. $l$/$w$/$h$ is the three box dimensions, and $\theta$ is box heading.
\section{Experiments}
\subsection{Datasets}
We conduct extensive experiments on three large-scale datasets to showcase the general applicability of FSDv2 after removing the inductive bias.
\subsubsection{Waymo Open Dataset}
Following previous FSDv1~\cite{fsd}, we first use Waymo Open Dataset~\cite{WOD} (WOD) to evaluate the performance of our proposed method.
With 1150 sequences and more than 200,000 frames, WOD is currently the largest dataset of its kind.
Among them, 798 sequences are used for training, 202 for validation, and 150 for testing.
The detection range in WOD is 75 meters (cover area of $150m \times 150m$).
\par
\subsubsection{Argoverse 2 Dataset}
Argoverse 2~\cite{argo2} (AV2) is a long-range dataset with a perception range of up to 200 meters, covering an area of $400m \times 400m$.
Such a large perception range makes dense/semi-dense detectors infeasible here. 
AV2 has a similar scale to WOD, containing 1000 sequences in total, 700 for training, 150 for validation, and 150 for testing.
In addition to distance-based average precision (AP), AV2 adopts a \textit{composite score} as an evaluation metric, which takes both AP and localization errors into account.
\par
Another distinctive characteristic of AV2
is the highly diverse object classes and sizes.
AV2 contains 30 annotated categories, ranging from common objects like cars and pedestrians to long-tailed objects such as strollers and wheelchairs.
Some objects reach up to 15 meters in length (e.g., buses) and some are as small as 0.1 meters (e.g., construction cones and bollards).
This diverse distribution of object classes and sizes makes AV2 an ideal testbed for evaluating detectors aiming for general applicability.
Due to so many many categories, we select the following representative categories to report the performance in some ablation studies.
\emph{Regular vehicle} and \emph{pedestrian}: they are the most common objects in autonomous driving scenes.
\emph{Bus}: it is the largest objects in AV2.
\emph{Construction Cone}, \emph{Bollard}, and \emph{Stop Sign}: they have very small sizes in Bird's Eye View.
\par
\subsubsection{nuScenes Dataset}
nuScenes dataset~\cite{nus} is another mainstream benchmark for LiDAR-based and image-based 3D object detection, containing 1000 scenes and 10 categories. It adopts distance-based average precision (AP) and \emph{NuScenes Detection Score} (NDS) as the evaluation metric.
NDS is mainly determined by AP but also takes pose estimation error into account.

\subsection{Setup}
\subsubsection{Model Variants}
\label{sec:variants}
Our implemented models slightly differ in different benchmarks.
On Waymo, FSDv2 inherits the second stage module from FSDv1 for object-centric refinement~\cite{pvrcnn, lidarrcnn, ba-det}.
On AV2 and nuScenes, we remove the second stage in the original FSD since it introduces more class-wise hyper-parameters hindering the model generalization, especially in datasets with multiple categories.
\par
We follow~\cite{centerpoint} to partition all categories into different tasks.
For WOD, each class is a single task as in~\cite{fsd, fsd++}.
For Argoverse 2, we follow the partition in the official baseline~\cite{argo2, fsd}.
For nuScenes, we use a 2-task partition. The first task contains Car, Truck, Construction Vehicle, Bus, and Trailer.
The other five categories are in another task.
We find such a partition is faster than the conventional 6-task partition~\cite{centerpoint} without performance loss in FSDv2, so we adopt it as the default setting.
We summarize the model variants in Table~\ref{tab:settings}.
These adjustments allow us to tailor the models to each benchmark while maintaining their core design principles.
\begin{table}[h]
	\begin{center}
		\resizebox{0.99\columnwidth}{!}{
			\begin{tabular}{lccc}
				\toprule
			
				 & WOD & AV2 & nuScenes  \\
				\midrule
				Two Stage & \cmark & \xmark & \xmark \\
                Task Head & 3 tasks & 6 tasks & 2 tasks\\
                Schedule & 24 & 24 & 24 (w/. CBGS~\cite{cbgs}) \\
                VV. size & $(0.5, 0.5, 0.5)$ & $(0.4, 0.4, 0.4)$ & $(0.4, 0.4, 0.4)$\\
                \#. Fading epochs & 3 & 5 & 5 \\
                $\alpha$ (in Eq.~\ref{eq:weighted_alpha}) & 1 & 0.5 & 1 \\
				 \bottomrule
			\end{tabular}
		}	
	\end{center}
		\caption{
  List of core hyper-parameters differing among three datasets.}
	\label{tab:settings}
\end{table}
\subsubsection{Implementation Details}
\noindent \textbf{Codebase and Devices.}
Our implementation is based on the official FSDv1 codebase, which is inherited from MMDetection3Dv0.15~\cite{mmdet3d} with PyTorch 1.8 and SpConv 2.2.3.
All experiments are conducted in 8 RTX 3090 GPUs.
\par
\noindent \textbf{Schedule and Optimization.}
Previous FSDv1 adopt a 12-epoch schedule.
Since we remove the handcrafted clustering, we find the performance of FSDv2 is less likely to saturate with longer training (analyzed in Sec.~\ref{sec:exp_longer_training}).
Thus we adopt a 24-epoch schedule for FSDv2 in all three datasets.
For nuScenes, we follow the convention to adopt CBGS~\cite{cbgs} for class-balance learning.
The hyper-parameters of the optimizer are the same as those in FSDv1.
For all datasets, we use batch size 2 on each GPU.
\par
\noindent \textbf{Data Augmentation.}
The data augmentations on WOD and Argoverse 2 are the same as those in FSDv1, including Global Flip, Global Rotation, and CopyPaste.
For nuScenes, we adopt the augmentation strategies in TransFusion-L~\cite{transfusion}.
We also adopt the fading strategy~\cite{pointaugmenting} to disable CopyPaste augmentation in the last few epochs.
Table~\ref{tab:settings} summarizes some important hyperparameters.
\par
\noindent More details can be found in our open-source code\footnote{\url{https://github.com/tusen-ai/SST}}.
\subsection{Main Results}
We present the main results in the three adopted datasets with standard settings in Table~\ref{tab:wod_validation}, Table~\ref{tab:wod_test}, Table~\ref{tab:argo_main}, and Table~\ref{tab:nus_main}.
FSDv2 achieves state-of-the-art performance in WOD and Argoverse 2 datasets and is on par with state-of-the-art methods in nuScenes.
Notably, FSDv2 surpasses previous fully sparse methods by a large margin in Argoverse 2.

\begin{table*}[ht]
\small
\centering

\resizebox{0.99\linewidth}{!}{
\begin{tabular}{l|c|c|c|c|c|c|c|c}
  \specialrule{1pt}{0pt}{1pt}
\toprule
\multirow{2}{*}{Methods}  & \multirow{2}{*}{\shortstack[1]{mAP/mAPH\\ L1}} & \multirow{2}{*}{\shortstack[1]{mAP/mAPH\\ L2}} & \multicolumn{2}{c|}{\textit{Vehicle} 3D AP/APH} & \multicolumn{2}{c|}{\textit{Pedestrian} 3D AP/APH} & \multicolumn{2}{c}{\textit{Cyclist} 3D AP/APH}\\
 & & &  L1      &   L2     &   L1      &   L2  &   L1     &   L2\\
\midrule

SECOND~\cite{second}& 67.2/63.1 &61.0/57.2 & 72.3/71.7 & 63.9/63.3 & 68.7/58.2 & 60.7/51.3 & 60.6/59.3 & 58.3/57.0\\


RangeDet~\cite{rangedet}& 71.5/69.5 &65.0/63.2 & 72.9/72.3 & 64.0/63.6 & {75.9}/71.9 & 67.6/63.9 & 65.7/64.4 & 63.3/62.1 \\

PointPillars~\cite{pointpillar} & 69.0/63.5 & 62.8/57.8 & 72.1/71.5  & 63.6/63.1 & 70.6/56.7  & 62.8/50.3 & 64.4/62.3 & 61.9/59.9 \\

LiDAR-RCNN~\cite{lidarrcnn} & 71.9/67.0 & 65.8/61.3 & 76.0/75.5 & 68.3/67.9 & 71.2/58.7 & 63.1/51.7 & 68.6/66.9 & 66.1/64.4\\

M3DETR~\cite{m3detr}& 68.7/65.2 &  61.8/58.7 & 75.7/75.1 & 66.6/66.0 & 65.0/56.4 & 56.0/48.4 & 65.4/64.2 & 62.7/61.5\\
Part-A2-Net~\cite{parta2} & 73.6/70.3 & 66.9/63.8 & 77.1/76.5 & 68.5/68.0 & 75.2/66.9 & 66.2/58.6 & 68.6/67.4 & 66.1/64.9\\
CenterPoint~\cite{centerpoint} & 75.9/73.5 & 69.8/67.6 & 76.6/76.0 & 68.9/68.4 & 79.0/73.4 & {71.0}/65.8 & 72.1/71.0 & 69.5/68.5\\
$\circ$ IA-SSD~\cite{iassd} & 69.2/64.5 & 62.3/58.1 & 70.5/69.7 & 61.6/61.0 & 69.4/58.5 & 60.3/50.7 & 67.7/65.3 & 65.0/62.7 \\

RSN ~\cite{rsn}& -/- &  -/- & 75.1/74.6 &66.0/65.5 & 77.8/72.7 & 68.3/63.7 & -/- & -/- \\
Voxel-RCNN~\cite{voxelrcnn} & 77.8/75.5 & 71.4/69.2 & 78.5/78.0 & 69.9/69.5 & 81.2/76.0 & 73.3/68.3 & 73.6/72.6 & 70.9/69.8 \\
SST\_TS~\cite{sst} & -/- & -/- & 76.2/75.8 & 68.0/67.6 & 81.4/74.0 & 72.8/{65.9} & -/- & -/-\\
AFDetV2~\cite{afdetv2} & 77.2/74.8 &  71.0/68.8 & {77.6/77.1} & {69.7}/69.2 & 80.2/74.6 & 72.2/67.0 & 73.7/72.7 & 71.0/70.1\\
PillarNet-34~\cite{pillarnet}& 77.4/74.6 & 71.0/68.5 & {79.1/78.6} &    {70.9}/   {70.5} & 80.6/74.0 & 72.3/66.2 & 72.3/71.2 & 69.7/68.7\\
PV-RCNN++(center)~\cite{pvrcnnpp} & 78.1/75.9 & 71.7/69.5 &     {79.3}/{78.8}   &  {70.6}/{70.2}&  81.3/76.3  &  73.2/68.0 & 73.7/72.7 & 71.2/70.2\\
ConQueR~\cite{conquer} & 79.4/77.0 & 74.0/71.6  & 78.4/77.9  & 71.0/70.5 & 82.4/76.6 & 75.8/70.1  & 77.5/76.4 & 75.2/74.1 \\
Graph-RCNN~\cite{graphrcnn} & 79.5/77.0 & 73.2/70.9 & \ul{80.8}/\ul{80.3} & \ul{72.6}/\ul{72.1} & 82.4/76.6 & 74.4/69.0 & 75.3/74.2 & 72.5/71.5 \\

GD-MAE~\cite{gdmae} & 80.2/77.6 &74.1/71.6 & 80.2/79.8 & \ul{72.4}/\ul{72.0} & 83.1/76.7 & 75.5/69.4 & 77.2/76.2 & 74.4/73.4  \\

CenterFormer~\cite{centerformer} & 75.6/73.2 &  71.4/69.1 & 75.0/74.4 & 69.9/69.4 & 78.0/72.4 & 73.1/67.7 & 73.8/72.7 & 71.3/70.2 \\
DSVT~\cite{dsvt} &\ul{81.1}/\ul{78.9} & \ul{74.8}/\ul{72.8}& \ul{80.4}/\ul{79.9} & 72.2/71.8 & \ul{84.2}/\ul{79.3} & \ul{76.5}/\ul{71.8} & \ul{78.6}/\ul{77.6} & \ul{75.7}/\ul{74.7}  \\
\midrule
$\circ$ FlatFormer~\cite{flatformer} & 79.3/77.1 & 72.7/70.5 & 78.6/78.1 & 69.8/69.4 & 82.9/77.5 & 74.3/69.3 & 76.6/75.6 & 73.9/72.8 \\
$\circ$ VoxelNeXt~\cite{voxelnext}  & 78.6/76.3 & 72.2/70.1 & 78.2/77.7 & 69.9/69.4 & 81.5/76.3 & 73.5/68.6 & 76.1/74.9 & 73.3/72.2 \\
$\circ$ FSDv1  & 79.6/77.4 &    {72.9}/{70.8} & 79.2/{78.8} & 70.5/{70.1} &    {82.6}/{77.3} &    {73.9}/{69.1} &    {77.1}/{76.0} &    {74.4}/{73.3} \\
$\circ$ FSDv2 (ours)  & \ul{81.8}/\ul{79.5} & \ul{75.6}/\ul{73.5}   & 79.8/79.3 & 71.4/71.0 & \ul{84.8}/\ul{79.7} & \ul{77.4}/\ul{72.5} & \ul{80.7}/\ul{79.6} & \ul{77.9}/\ul{76.8}  \\

\bottomrule
  \specialrule{1pt}{1pt}{0pt}
\end{tabular}
}
\vspace{2mm}
\caption{
    Comparison with the state-of-the-art detectors in Waymo Open Dataset \textbf{validation} split with the standard single-frame single-model setting.
    We \ul{underline} the best and second-best entries.
    }
    \vspace{-3mm}
 \label{tab:wod_validation}

\end{table*}
\begin{table*}[ht]
\small
\centering

\resizebox{0.99\linewidth}{!}{
\begin{tabular}{l|c|c|c|c|c|c|c|c}
  \specialrule{1pt}{0pt}{1pt}
\toprule
\multirow{2}{*}{Methods} & \multirow{2}{*}{\shortstack[1]{mAP/mAPH\\ L1}} & \multirow{2}{*}{\shortstack[1]{mAP/mAPH\\ L2}} & \multicolumn{2}{c|}{\textit{Vehicle} 3D AP/APH} & \multicolumn{2}{c|}{\textit{Pedestrian} 3D AP/APH} & \multicolumn{2}{c}{\textit{Cyclist} 3D AP/APH}\\
 & &&   L1      &   L2     &   L1      &   L2  &   L1     &   L2\\
\midrule

CenterPoint~\cite{centerpoint}& -/- & -/69.0 & -/- & -/71.9 & -/- & -/67.0 & -/- & -/68.2\\
AFDetV2-lite~\cite{afdetv2}& 77.6/75.2 & 72.2/70.3 & 80.5/80.0 & 73.0/72.6 & 79.8/74.3 & 73.7/68.6 & 72.4/71.2 & 69.8/69.7 \\
PV-RCNN~\cite{pvrcnn}& 76.9/74.2  & 71.3/68.8 & 80.6/80.1     & 72.8/72.4 & 78.2/72.0  &  71.8/66.0 & 71.8/70.4 & 69.1/67.8\\
PV-RCNN++~\cite{pvrcnnpp}& 78.0/75.7  & 72.4/70.2 & 81.6/81.2    & 73.9/73.5 &  80.4/75.0  &  74.1/69.0 & 71.9/70.8 & 69.3/68.2\\
GD-MAE~\cite{gdmae} & 80.3/77.7 & 74.7/72.3 & 83.6/83.2 &	75.8/75.5 & 83.2/77.1 &	77.1/71.3 & 74.0/72.9 &	71.2/	70.2 \\
Graph-RCNN~\cite{graphrcnn} & 79.3/77.0 & 73.8/71.6 & 83.6/83.1 &	76.0/75.6 & 81.9/76.5 &	75.6/70.5 & 72.5/71.3 &	69.8/68.7 \\
$\circ$ FSDv1~\cite{fsd} & 80.4/78.2 & {74.4}/{72.4} & {82.7}/{82.3} & {74.4}/{74.1} & {82.9}/{77.9} & {75.9}/{71.3} & {75.6}/{74.4} & {72.9}/{71.8} \\
$\circ$ FSDv2 (Ours) & 81.1/79.0 & 75.4/73.3 & 82.4/82.0 &	74.4/74.0 & 83.8/78.9 &	77.4/72.8 & 77.1/76.0 & 74.3/73.2\\
\bottomrule
  \specialrule{1pt}{1pt}{0pt}
\end{tabular}
}
\vspace{2mm}
\caption{
Comparison with the state-of-the-art detectors in Waymo Open Dataset \textbf{test} split with the standard single-frame single-model setting.
    }
    \vspace{-3mm}
 \label{tab:wod_test}

\end{table*}

\begin{table*}[ht]
\setlength{\tabcolsep}{2pt}

\begin{center}
\resizebox{\textwidth}{!}{%
\begin{tabular}{l|c|cccccccccccccccccccccccccccccc}
\toprule
  {Methods} &
 \rotatebox{90} {{mAP}} & 
 \rotatebox{90} {Vehicle} & 
 \rotatebox{90} {Bus} &
 \rotatebox{90} {{Pedestrian}} &
 \rotatebox{90} {{Stop Sign}} &
 \rotatebox{90} {Box Truck} &
 \rotatebox{90} {Bollard} &
 \rotatebox{90} {{C-Barrel}} &
 \rotatebox{90} {{Motorcyclist}} &
 \rotatebox{90} {MPC-Sign} &
 \rotatebox{90} {{Motorcycle}} &
 \rotatebox{90} {Bicycle} &
 \rotatebox{90} {{A-Bus}} &
 \rotatebox{90} {{School Bus}} &
 \rotatebox{90} {Truck Cab} &
 \rotatebox{90} {{C-Cone}} &
 \rotatebox{90} {V-Trailer} &
 \rotatebox{90} {Sign} &
 \rotatebox{90} {Large Vehicle} &
 \rotatebox{90} {{Stroller}} &
 \rotatebox{90} {{Bicyclist}} &
  \rotatebox{90} {Truck} &
  \rotatebox{90} {MBT} &
    \rotatebox{90} {Dog} &
        \rotatebox{90} {Wheelchair} &
           \rotatebox{90} {W-Device} &
             \rotatebox{90} {W-Rider} &

\\ 
\midrule
CenterPoint\dag~\cite{centerpoint}        & 13.5 & 61.0 & 36.0  & 33.0  & 28.0  & 26.0  & 25.0  & 22.5  & 16.0   & 16.0 & 12.5 & 9.5  & 8.5 & 7.5 & 8.0 & 8.0  & 7.0 & 6.5  &3.0 & 2.0 & 14.0 & 14.0 & 1.0 & 0.5 & 0 & 3.0 & 0 \\
CenterPoint$^\ast$       & 22.0 & 67.6 & 38.9 & 46.5 & 16.9 & 37.4 & 40.1 & 32.2 & 28.6 & 27.4 & 33.4 & 24.5 & 8.7 & 25.8 & 22.6 & 29.5  & 22.4 & 6.3 & 3.9 & 0.5 & 20.1 & 22.1 & 0 & 3.9 & 0.5 & 10.9 & 4.2\\
$\circ$ FSDv1      & 28.2 & 68.1 & 40.9 & 59.0 & 29.0 & 38.5 & 41.8 & 42.6 & 39.7 & 26.2 & 49.0 & 38.6 & 20.4 & 30.5 & 14.8 & 41.2  & 26.9 & 11.9 & 5.9 & 13.8 & 33.4 & 21.1 & 0 & 9.5 & 7.1 & 14.0 & 9.2\\
$\circ$ VoxelNeXt & 30.7 & 72.7 & 38.8 & 63.2 & 40.2 & 40.1 & 53.9& 64.9& 44.7 & 39.4 & 42.4 & 40.6 & 20.1 & 25.2 & 19.9 & 44.9 & 20.9 & 14.9 & 6.8 & 15.7 & 32.4 & 16.9 & 0 & 14.4 & 0.1 & 17.4 & 6.6 \\
$\circ$ FSDv2 w/o. CP \ddag & 36.0 & 76.6 & 46.0 & 69.8 & 39.9 & 39.5 & 52.8 & 57.6 & 55.4 & 35.4 & 57.8 & 49.2 & 24.7 & 37.2 & 27.4 & 41.3 & 32.1 & 16.5 & 6.4 & 36.1 & 44.6 & 22.5 & 0.9 & 12.1 & 14.7 & 23.3 & 16.0 \\
$\circ$ FSDv2 & 37.6 & 77.0 & 47.6 & 70.5 & 43.6 & 41.5 & 53.9 & 58.5 & 56.8 & 39.0 & 60.7 & 49.4 & 28.4 & 41.9 & 30.2 & 44.9 & 33.4 & 16.6 & 7.3 & 32.5 & 45.9 & 24.0 & 1.0 & 12.6 & 17.1 & 26.3 & 17.2 \\
\bottomrule

\end{tabular}%
}
\end{center}
\caption{
{Performance in Argoverse 2 validation split.}
\textbf{\dag}: provided by authors of AV2 dataset.
$\ast$: re-implemented by ourselves.
\textbf{\ddag}: VoxelNeXt observed that disabling CopyPaste augmentation is better, so we also disable this augmentation for fair comparison.
C-Barrel: construction barrel.
MPC-Sign: mobile pedestrian crossing sign.
A-Bus: articulated bus.
C-Cone: construction cone.
V-Trailer: vehicular trailer.
MBT: message board trailer.
W-Device: wheeled device.
W-Rider: wheeled rider.
$\circ$ indicates fully sparse detectors.
}
\vspace{-3mm}
\label{tab:argo_main}
\end{table*}

\begin{table*}[htbp]
\begin{center}

\resizebox{0.97\textwidth}{!}{%
\begin{tabular}{l|cc|cc|ccccccccccc}
  \specialrule{1pt}{0pt}{1pt}
\toprule
\multirow{2}{*}{Method}  & \multicolumn{2}{c|}{\textit{val}} & \multicolumn{12}{c}{\textit{test}}                                        \\ 
    & mAP   &    NDS  & mAP &     NDS & Car  & Truck   & Bus   & Trailer  & C.V. & Ped. & Mot. & Byc. & T.C. & Bar.  \\
                        \midrule
CenterPoint~\cite{centerpoint}                           & 59.6        &  66.8       & 60.3       &  67.3  & 85.2 & 53.5 & 63.6 & 56.0 & 20.0 & 84.6 & 59.5 & 30.7 & 78.4 & 71.1   \\
CBGS~\cite{cbgs} &51.4 & 62.6 & 52.8 & 63.3& 81.1 & 48.5 & 54.9 & 42.9 & 10.5  & 80.1 & 51.5 & 22.3 & 70.9 & 65.7\\
CVCNET~\cite{cvcnet} &54.6& - & 55.8 & 64.2 & 82.7 & 46.1 & 45.8 & 46.7 & 20.7 & 81.0 & 61.3 & 34.3 & 69.7 & 69.9\\
HotSpotNet~\cite{hotspotnet} & 59.5 & 66.0 &59.3&66.0& 83.1 & 50.9&  56.4&  53.3&  23.0&  81.3&  63.5&  36.6&  73.0&  71.6\\
AFDetV2~\cite{afdetv2} & - & - &62.4&68.5& 86.3& 54.2& 62.5& 58.9& 26.7& 85.8& 63.8& 34.3& 80.1& 71.0\\
VISTA~\cite{vista}\dag & 60.8 & 68.1 &63.0 & 69.8 & 84.4& 55.1& 63.7& 54.2& 25.1& 82.8& 70.0& 45.4& 78.5& 71.4 \\
UVTR-L~\cite{uvtr}\dag & 60.9 &67.7 &63.9& 69.7& 86.3& 52.2& 62.8& 59.7 &33.7& 84.5& 68.8& 41.1& 74.7& 74.9\\
PillarNet-18~\cite{pillarnet}\dag & 59.9 & 67.4 &65.5 & 70.0 & 87.4& 56.7 &60.9 &61.8& 30.4& 87.2& 67.4& 40.3& 82.1& 76.0 \\
Focals Conv~\cite{focalspconv} &61.2 & 68.1 &63.8 &70.0& 86.7 &56.3& 67.7& 59.5& 23.8& 87.5& 64.5& 36.3& 81.4& 74.1        \\
TransFusion-L~\cite{transfusion}                        & 65.1        &  70.1       & 65.5       &   70.2   & 86.2 & 56.7 & 66.3& 58.8& 28.2& 86.1& 68.3& 44.2& 82.0& 78.2    \\
LargeKernel3D~\cite{largekernel3d}                             & 63.3        &  69.1       & 65.4       &   70.6  & 85.5& 53.8& 64.4& 59.5& 29.7& 85.9& 72.7& 46.8& 79.9& 75.5   \\
\midrule
$\circ$ FSDv1~\cite{fsd}                                & 62.5        &  68.7       & -          &   - & -&-&-&-&- &- &-& - &- &-\\
$\circ$ VoxelNeXt~\cite{voxelnext}                               & 63.5        &  68.7       & 64.5          &   70.0  & 84.6& 53.0& 64.7& 55.8& 28.7& 85.8& 73.2& 45.7& 79.0 &74.6   \\
$\circ$ FSDv2 & 64.7 & 70.4 & 66.2 & 71.7 & 83.7 & 51.6 & 66.4 & 59.1 & 32.5 & 87.1 & 71.4 & 51.7 & 80.3 & 78.7\\
\bottomrule
  \specialrule{1pt}{1pt}{0pt}
\end{tabular}%
}
\end{center}
\caption{
{Performance on the nuScenes dataset.} 
We do not use test-time augmentation or model ensemble.
\dag: using test-time augmentation in test split.
C.V.: construction vehicle.
Ped.: pedestrian.
Mot.: motorcyclist.
Byc.: bicyclist.
T.C.: traffic cone.
Bar.: barrier.
$\circ$: indicates fully sparse methods. 
}
\vspace{-3mm}
\label{tab:nus_main}
\end{table*}

\subsection{Cluster v.s. Virtual Voxel}
To gain a deeper understanding of the differences between clustering in FSDv1 and the proposed virtual voxel mechanism, we compare them from various perspectives.
\subsubsection{Statistics}
We begin by analyzing the distribution of clusters and virtual voxels within an object.
Table~\ref{tab:stats} shows the average number of point groups (i.e., clusters or virtual voxels) of each ground-truth object. We ignore false negative points.
A couple of findings can be concluded as follows.
\begin{itemize}
    \item For large objects, such as vehicles, there are usually several virtual voxels associated with each object.
    The number of virtual voxels tends to be larger than the number of clusters, necessitating the use of a virtual voxel mixer to aggregate their features.
    \item For small objects like \emph{Construction Cone} and \emph{Bollard}, each object contains very few voxels or clusters. In such cases, virtual voxels and clusters have similar functionality.
    \item The number of \textbf{real} voxels of large \emph{Vehicle} is much larger than the number of clusters and virtual voxels.
    It suggests that making predictions from all real voxels may lead to a severe imbalance between objects of different sizes.
    Instead, making predictions from a smaller number of virtual voxels potentially mitigates the problem.
\end{itemize}
\begin{table}[H]
	\begin{center}
		\resizebox{0.85\columnwidth}{!}{
			\begin{tabular}{l|cccc}
				\toprule
			
				Group Type & Vehicle & Pedestrian & C.C & Bollard \\
				\midrule
    				\#. clusters & 1.2  & 1.3 & 0.91$^\ast$ & 0.83\\
				\#. virtual voxels & 11.4 & 2.4 & 0.97 & 0.92\\
                    \#. real voxels & 37.7 & 6.24 & 1.2 &  1.1\\
		 \bottomrule
			\end{tabular}
		}	
	\end{center}
		\caption{Statistics of virtual voxels in FSDv2 and clusters in FSDv1. The statistics are obtained from well-trained detectors on Argoverse 2 with a virtual voxel size of $[0.4m \times 0.4m \times 0.4m]$.
        $^\ast$: There are numbers less than 1 because some objects are recognized as background by detectors, while we use the number of total ground-truth objects as the divisor.
  }
  \vspace{-3mm}
	\label{tab:stats}
\end{table}

\subsubsection{Performance in crowded scenes}
One drawback of instance-level representation in FSDv1 is that multiple instances in crowded scenes can be mistakenly recognized as a single instance.
To quantitatively analyze this issue, we define an object in a crowded scene if the distance between the object and its nearest neighbor object in the same category is smaller than 2 meters.
Finally, we obtain two quantitative findings as follows.
\begin{itemize}
    \item In FSDv1, approximately 6.9\% of crowded pedestrians are completely clustered together with others\footnote{There are also pedestrians partially clustered together with others.}, resulting in considerable false negatives.
    \item We further evaluate the detection performance using crowdedness as a breakdown, Table \ref{tab:crowd} shows the results. As observed, FSDv2 gains much more in crowded cases than in normal cases, showcasing the benefits of removing handcrafted instance representation.
\end{itemize}

\begin{table}[H]
\setlength{\tabcolsep}{2pt}
	\begin{center}
		\resizebox{0.99\columnwidth}{!}{
			\begin{tabular}{l|ccc|ccc}
				\toprule
			
				 &\multicolumn{3}{c|}{L2 AP} & \multicolumn{3}{c}{L2 AR} \\
				& Normal & Crowded & Overall & Normal & Crowded & Overall \\
				\midrule
				FSDv1  & 75.1 & 71.9 & 74.4 & 80.2 & 75.9 & 78.2\\
				FSDv2  & 76.8 (+1.7) & 76.7 (+\textbf{4.8}) & 77.4  & 85.4 (+5.2) & 84.6 (+\textbf{8.7}) & 85.4  \\

				 \bottomrule
			\end{tabular}
		}	
	\end{center}
		\caption{Performance on Waymo Open Dataset under crowdedness breakdown.}
  \vspace{-3mm}
	\label{tab:crowd}
\end{table}

\subsubsection{Effectiveness of eliminating inductive bias.}
\label{sec:exp_longer_training}
\begin{figure}[H]
	\centering
	\includegraphics[width=0.9\linewidth]{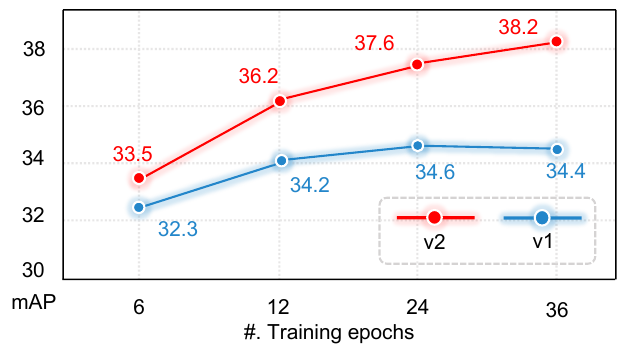}
  \vspace{-4mm}
	\caption{\textbf{Performance w.r.t. training epochs}. Experiments are conducted on Argoverse 2 dataset. FSDv2 consistently outperforms the previous FSDv1 and showcases even better performance with longer training.}
	\label{fig:epochs}
\end{figure}

Without the inductive bias introduced by handcrafted clustering, the model gains the potential for better performance with longer train schedules.
So we conduct experiments with different training schedules on FSDv1 and FSDv2.
Fig.~\ref{fig:epochs} shows the performance trend as the training epoch increases.
It is a clear trend that the performance of FSDv1 saturates very soon,
reaching a plateau at epoch 12.
In contrast, FSDv2 constantly exhibits better performance as the epochs increase.

\begin{table*}[htbp]
\setlength{\tabcolsep}{3pt}
\begin{center}

\resizebox{0.99\textwidth}{!}{%
\begin{tabular}{l|cccc|cccccccc|cccccccc}
  \specialrule{1pt}{0pt}{1pt}
\toprule
\multirow{2}{*}{Model} & \multicolumn{4}{c|}{Waymo} & \multicolumn{8}{c|}{Argoverse 2} &  \multicolumn{8}{c}{nuScenes}                                       \\ 
     &       Mean           & Veh.   &    Ped. & Cyc. &    mAP & CDS   & Veh.  & Bus & Ped. & Bol. & C.C. & S. Sign & mAP & NDS  & Car & Truck & Bus & Ped. & T.C. & Bar. \\
                        \midrule

FSDv2-Baseline  & 72.8 & 70.1 & 71.7 & 76.6 & 35.4 & 28.0 & 73.9 & 45.1 & 68.2 & 48.7 & 40.8 & 35.9 & 60.9 & 67.5 & 81.1 & 54.0 & 71.4 & 86.0 & 69.6 & 66.1 \\
Above + Mixer$^\ast$ & 73.0 & 70.3 & 72.8\red{$\uparrow$} & 76.1 & 36.2 & 28.7 & 74.9 & 44.7 & 68.0 & 50.1 & 40.9 & 36.9 & {62.2}\red{$\uparrow$} & {68.5}\red{$\uparrow$} & 82.2 & {56.6}\red{$\uparrow$} & 72.3 & 85.4 & 69.7 & {69.2}\red{$\uparrow$}\\
Above + VV\dag & 73.2 & 70.8\red{$\uparrow$} & 72.7 & 76.2 & 36.7 & 28.9 & {76.4}\red{$\uparrow$} & {47.4}\red{$\uparrow$} & 68.7 & 51.4 & 42.8 & 39.7 & 62.7 & 68.8 & {83.5}\red{$\uparrow$} & 56.6 & {73.9}\red{$\uparrow$} & 85.6 & 69.8 & {72.7}\red{$\uparrow$}\\
Above w/o. VA\ddag & 73.0 & 70.8 & 72.6 & 75.7 & 34.3 & 27.5 & 76.6 & 46.7 & 67.5 & {34.1}\dgreen{$\downarrow$} & {30.1}\dgreen{$\downarrow$} & {33.8}\dgreen{$\downarrow$} & 61.9 & 68.6 & 83.4 & 56.5 & 74.6 & 85.1 & {64.3}\dgreen{$\downarrow$} & 71.1\\
\bottomrule
  \specialrule{1pt}{1pt}{0pt}
\end{tabular}%
}
\end{center}
\caption{
{Overall ablations of main components.}
We mark the numbers which have significant changes from their previous row with \red{$\uparrow$} and \dgreen{$\downarrow$}.
Explanations of the variants:
\textbf{(1)} $^\ast$: Incorporating virtual voxel mixer, but it only processes the real voxels, playing a simple rule of feature enhancement.
\textbf{(2)} \dag: Further incorporating virtual voxelization, which is actually the default setting of FSDv2.
\textbf{(3)} \ddag: Replacing the weighted centroids in virtual voxel assignment with geometric centers.
Cyc.: cyclist.
Bol.: bollard.
C.C.: construction cone.
S. Sign: stop sign.
Please refer to Table~\ref{tab:nus_main} for other abbreviations.
}
\vspace{-3mm}
\label{tab:baselines}
\end{table*}
\subsection{Main Ablations}
In this section, we start by establishing a clean baseline and then gradually incorporate the proposed components one by one to showcase their individual effectiveness and the overall performance roadmap.
All experiments in this section adopt a 12-epoch schedule, which greatly shortens the experimental period without significant performance loss.
\par

\subsubsection{Baseline Setup}
We refer to the baseline model as FSDv2-baseline, which removes all the components related to the virtual voxel mechanism.
The specific characteristics of the FSDv2-baseline are as follows.
\begin{itemize}
    \item FSDv2-baseline does not generate voted centers, but we retain the voting loss as an auxiliary supervision for a fair comparison.
    \item We still employ voxelization and voxel encoding after center voting. However, here they deal with real points without incorporating the voted centers. So there are no virtual voxels anymore.
    \item The encoded real voxels are then directly sent to the detection head without passing through the mixer. And predictions are made from these real voxels.
    \item Note that we find virtual voxel assignment has a significant impact on the performance of small objects (Table~\ref{tab:baselines}). So we keep it in FSDv2-baseline in case of reaching inaccurate conclusions.
\end{itemize}
Starting from the baseline model, we incrementally add proposed components to reveal the performance trend.
Table~\ref{tab:baselines} shows the results of the overall ablations, and list several conclusions as follows.
\begin{itemize}
    \item Across all three datasets, our designs consistently lead to performance improvements, demonstrating the effectiveness and general applicability of our designs.
    \item While not entirely absolute, the overall trend suggests that the proposed mixer and virtual voxels enhance the performance of objects with relatively large sizes (e.g., Vehicle, Bus).
    \item Notably, the virtual voxel assignment plays a crucial role in boosting the performance of small objects, such as Bollards, Construction/Traffic Cones, and Stop Signs.
\end{itemize}

\subsection{Performance Analysis of Each Component}
In this subsection, we delve into the detailed analysis of each component for a better understanding of the workings.
\subsubsection{Effectiveness of Virtual Voxel}
To understand the effect of the virtual voxel mechanism, we design a degradation strategy to progressively degrade the effect of virtual voxels.
In particular, we introduce a continuous scaling factor $s \in [0, 1]$ to multiply with the predicted center voting offsets.
In default FSDv2, $s$ is set to 1, fully preserving the effect of virtual voxels. As we decrease $s$ towards 0, the impact of virtual voxels is gradually diminished.
The results of different scaling factors are listed in Table~\ref{tab:virtual_voxel_scaling}.
We have two findings from Table~\ref{tab:virtual_voxel_scaling}.
\begin{itemize}
    \item Large objects have better performance with larger scaling factors. Since larger objects are more likely to have empty centers, it is reasonable that the virtual voxels have a significant effect on them. 
    \item In contrast, the scaling factor has minimal effect on small objects. This is reasonable because, for small objects, the voting offsets are also small, virtual voxels with varying scaling factors occupy similar spatial positions.
    \item The performance of Barrier gets greatly boosted with a larger factor. We postulate the following: barriers are typically placed in close proximity to one another.
    With small $s$, the detector tends to make predictions from the boundary between two adjacent barriers, which may introduce ambiguities. Instead, making predictions from virtual voxels located closer to the centers mitigates such ambiguities.
\end{itemize}

\begin{table}[H]
\setlength{\tabcolsep}{4pt}
	\begin{center}
		\resizebox{0.99\columnwidth}{!}{
			\begin{tabular}{l|cccccccc}
				\toprule
			
				{Voting scaling} & mAP & NDS & Car & Truck & Bus & Ped. & T.C. & Bar. \\
				\midrule
				0.0 & 61.3 & 68.2 & 82.0 & 55.9 & 72.8 & 85.3 & 69.4 & 66.8\\
				0.25 & 61.9 & 68.4 & 83.0 & 55.7 & 73.2 & 85.4 & 69.6 & 70.2\\
				0.5 & 62.4 & 68.7  & 83.2 & 56.0 & 73.4 & 85.4 & 69.7 & 70.9\\
    			0.75 & 62.6 & 68.9 & 83.2 & 56.3 & 73.5 & 85.6 & 69.6 & 71.4 \\
    			1.0 & 62.7 & 68.8 & 83.5 & 56.6 & 73.9 & 85.6 & 69.8 & 72.7 \\

				 \bottomrule
			\end{tabular}
		}	
	\end{center}
		\caption{{Tuning the effect of virtual voxels on nuScenes.} The voting scaling factor controls the influence of virtual voxels. A factor of 1 fully preserves the effect of virtual voxels, while a factor of 0 eliminates the effect.}
	\label{tab:virtual_voxel_scaling}
\end{table}

\subsubsection{Effectiveness of Virtual Voxel Encoder (VVE)}
As shown in Fig.~\ref{fig:voting_cases}, encoding the virtual voxel aggregates the features of object parts or the complete object. 
Although the virtual voxel mixer also performs feature aggregation, we empirically find VVE could help more.
To demonstrate this, we conduct experiments with zeroing the features of voted centers, while the features of real voxels are still there.
Thus, the features of virtual voxels can be only aggregated from real voxels by the virtual voxel mixer.
Table~\ref{tab:vve} shows the results, where VVE has a larger impact on small objects.
This observation can be explained by the fact that a virtual voxel often covers a significant portion of a small object.
\begin{table}[H]
\setlength{\tabcolsep}{4pt}
	\begin{center}
		\resizebox{0.9\columnwidth}{!}{
			\begin{tabular}{l|cccccccc}
				\toprule
			
				{VVE} & mAP & NDS & Car & Truck & Bus & Ped. & T.C. & Bar.\\
				\midrule
				\cmark & 62.7 & 68.8 & 83.5 & 56.6 & 73.9 & 85.6 & 69.8 & 72.7\\
				 \xmark  & 62.1 & 68.4 & 83.3 & 56.6 & 73.7 & 85.0 & 69.1 & 69.9\\
				 \bottomrule
			\end{tabular}
		}	
	\end{center}
		\caption{The effectiveness of Virtual Voxel Encoder (nuScenes).}
	\label{tab:vve}
\end{table}

\subsubsection{Input of Virtual Voxel Mixer}
\label{sec:exp_mixer}
In Sec.~\ref{sec:mixer}, we propose three kinds of input for the virtual voxel mixer. (1) Only taking virtual voxel features as input.
(2) Taking both virtual voxel features and real voxel features as input. (3) Taking virtual voxel features and multi-scale real voxel features as input. The results in Table~\ref{tab:mixer} suggest that incorporating real voxels significantly improves the performance, which is a notable difference from the FSDv1.
In FSDv1, background points are discarded after the instance segmentation, leading to potential false negatives and information loss. 
However, the inclusion of real voxels also comes with more cost, which is analyzed in Sec.~\ref{sec:runtime}, and we find the overhead is acceptable.
 
\begin{table}[H]
\setlength{\tabcolsep}{2pt}
	\begin{center}
		\resizebox{0.99\columnwidth}{!}{
			\begin{tabular}{l|cccccccc}
				\toprule

				& mAP & NDS & Car & Truck & Bus & Ped. & T.C. & Bar.\\
				\midrule
				Virtual only & 60.6  & 67.4 & 79.8 & 52.8 & 70.7 &84.6 & 69.5 & 70.5\\
				Virtual + Real & 62.0 & 68.5 & 82.8 & 56.7 & 74.2 & 85.6 & 69.7 & 71.1  \\
				Virtual + MS. Real & 62.7 & 68.8 & 83.5 & 56.6 & 73.9 & 85.6 & 69.8 & 72.7    \\

				 \bottomrule
			\end{tabular}
		}	
	\end{center}
		\caption{\textbf{Different inputs of virtual voxel mixer.} ``Real'' means real voxels. ``MS. Real'' means multi-scale real voxels. Evaluated on nuScenes.}
	\label{tab:mixer}
\end{table}

\subsubsection{Effectiveness of Virtual Voxel Assignment}
In Sec.~\ref{sec:assignment}, we claim that using the weighted centroid as the ``voxel position'' for assignment is much more friendly to tiny objects than using the geometric voxel center.
Moreover, some trivial solutions like enlarging the ground-truth bounding boxes also have the potential to deal with the label assignment of tiny objects.
To gain more insights, we conduct experiments on the Argoverse 2 dataset since it contains more tiny objects than the nuScenes dataset.
We draw the following conclusions from Table~\ref{tab:assignment}.
\begin{itemize}
    \item Slightly enlarging GT bounding boxes help assign more labels to tiny objects. However, aggressive enlarging is detrimental to the performance, which causes ambiguity when incorporating too much background clutter or two boxes overlapping.
    \item Using a virtual voxel centroid as its position is much better than the geometric center for tiny objects. And further using weighted centroids helps more.
\end{itemize}

\begin{table}[H]
\setlength{\tabcolsep}{3pt}
	\begin{center}
		\resizebox{0.99\columnwidth}{!}{
			\begin{tabular}{l|cccccccc}
				\toprule

				Strategies & mAP & CDS & Veh. & Bus & Ped. & Bol. & C.C. & S.Sign \\
				\midrule
				Geo. Center & 34.3 & 27.5 & 76.6 & 46.7 & 67.5 & 34.1 & 30.1 & 33.8\\
				Enlarge GT (0.2m) & 34.6 & 27.7 & 76.5 & 46.9 & 67.8 & 37.0 & 34.2 & 34.8\\
    			Enlarge GT (0.5m) & 33.5 & 26.9 & 76.1 & 46.0 & 66.0 & 32.6 & 28.7 & 31.3\\
				Centroid  & 36.2 & 28.9 & 76.4 & 47.4 & 68.7 & 51.4 & 42.8 & 39.7\\
				Weighted ($\alpha = 0.5$) & 36.2 & 28.9 & 76.6 & 46.8 & 69.1 & 52.5 & 43.0 & 39.1  \\
                    Weighted ($\alpha = 0.0$) & 36.9 & 29.5 & 76.6 & 47.2 & 69.4 & 52.4 & 43.7 & 39.8\\

				 \bottomrule
			\end{tabular}
		}	
	\end{center}
		\caption{{Different designs in the label assignment for virtual voxels.} Geo. Center: geometric center (Fig.~\ref{fig:assignment} (a)). Enlarging GT by $m$ meters means that each side of the GT bounding box is enlarged by $m$ meters. }
	\label{tab:assignment}
\end{table}

\par
The virtual voxel assignment treats all virtual voxels inside boxes as positive.
Recent methods~\cite{rsn, voxelnext} propose to only treat the nearest voxel to the object center as positive.
We argue this strategy is sub-optimal in Sec.~\ref{sec:potential_assignments}.
To validate this, we compare our strategy with the nearest assignment. The results are listed in Table~\ref{tab:nearest_assignment}.
As observed, the nearest assignment leads to a significant performance drop in large objects.
Incorporating more voxels as positive mitigates the performance loss.
In particular, performance saturates with 10 nearest virtual voxels since they account for a consideration portion of total virtual voxels as Table~\ref{tab:stats} shows.

\begin{table}[H]
\setlength{\tabcolsep}{3pt}
	\begin{center}
		\resizebox{0.99\columnwidth}{!}{
			\begin{tabular}{l|cccccccc}
				\toprule

				Strategies & mAP & CDS & Veh. & Bus & Ped. & Bol. & C.C. & S.Sign \\
				\midrule
				Ours (all) & 36.2 & 28.9 & 76.6 & 46.8 & 69.1 & 52.5 & 43.0 & 39.1  \\
    		      Nearest (top1) & 35.0 & 28.0 & 73.0\down & 41.7\down & 68.3 & 52.0 & 42.9 & 38.0  \\
				Top5 & 35.3 & 28.2 & 75.1\down & 43.0\down & 69.2 & 52.4 & 43.2 & 39.0  \\
    			Top10 & 36.1 & 28.9 & 76.4 & 45.9 & 69.1 & 52.4 & 42.9 & 39.2  \\

				 \bottomrule
			\end{tabular}
		}	
	\end{center}
		\caption{Comparison with other label assignment strategies.
  We mark the significant performance drop compared with first row with \down{}.
  }
	\label{tab:nearest_assignment}
\end{table}

\subsubsection{Virtual Voxel Size}
Unlike common voxelization in the early stages, virtual voxelization does not necessitate high resolution since features have been well-encoded by the backbone.
Thus we prefer relatively larger virtual voxel sizes for efficiency.
The results in Table~\ref{tab:virtual_voxel_size} suggest the performance is overall robust to the virtual voxel sizes.
In addition, larger voxel sizes lead to slightly better performance.
This can be attributed to the fact that a larger virtual voxel size leads to fewer virtual voxels on large objects, thereby mitigating the imbalance between objects of different sizes.

\begin{table}[H]
\setlength{\tabcolsep}{3pt}
	\begin{center}
		\resizebox{0.99\columnwidth}{!}{
			\begin{tabular}{l|cccccccc}
				\toprule
			
				Sizes (m) & mAP & NDS & Car & Truck & Bus & Ped. & T.C. & Bar \\
				\midrule
				(0.4, 0.4, 0.4) & 62.0 & 68.5 & 82.8 & 56.7 & 74.2 & 85.6 & 69.7 & 71.1\\
				(0.3, 0.3, 0.4) & 61.7 & 68.2 & {82.2}\down & {56.1}\down & {72.9}\down & 85.5 & 69.9 & 71.3 \\
    			(0.5, 0.5, 0.4) & 61.9 & 68.2 & 82.9 & 56.5 & 74.1 & 85.1 & {69.2}\down & 71.2 \\
				(0.4, 0.4, 0.3) & 61.8 & 68.5 & {82.3}\down & 56.5 & {73.1}\down & 85.6 & 69.8 & 70.9  \\
                    (0.4, 0.4, 0.5) & 62.0 & 68.6 & 82.7 & 56.7 & 74.1 & 85.5 & 69.2 & 71.2 \\

				 \bottomrule
			\end{tabular}
		}	
	\end{center}
		\caption{{Performance with different virtual voxel sizes.} We use \down{} to denote some entries which have notable performance drop compared with the first row.}
	\label{tab:virtual_voxel_size}
\end{table}

\subsubsection{Additional Techniques on Argoverse 2}
In addition to the proposed components, we make some other modifications to FSDv1 on Argoverse 2.
This is because previous FSDv1 adopts different settings on WOD and Argoverse 2 datasets, while we align the core settings across all three datasets.
In particular, on Argoverse 2, FSDv2 adopts a larger backbone, a longer training scheme, and better CopyPaste augmentations. Thus we show the effectiveness of these techniques in Table~\ref{tab:argo_roadmap} for a fair comparison.

\begin{table}[H]
\setlength{\tabcolsep}{3pt}
	\begin{center}
		\resizebox{0.99\columnwidth}{!}{
			\begin{tabular}{l|cccccccc}
				\toprule

				Strategies & mAP & CDS & Veh. & Bus & Ped. & Bol. & C.C. & S.Sign \\
				\midrule
				FSDv1 & 28.2 & 22.7 & 68.1 & 40.9 & 59.0 & 41.8 & 41.2 & 29.0\\
				FSDv2$^\ast$ & 29.8 & 23.7 & 71.9 & 43.4 & 63.7 & 44.2 & 37.0 & 25.1  \\
				FSDv2 (larger)\dag & 32.6 & 25.9 & 76.4 & 45.5 & 67.7 & 50.3 & 39.8 & 35.8  \\
                    FSDv2 w/. fade \ddag & 36.2 & 28.9 & 76.4 & 47.4 & 68.7 & 51.4 & 42.8 & 39.7 \\
                    FSDv2 $2\times$ & 37.6 & 30.2 & 77.0 & 47.6 & 70.5 & 53.9 & 44.9 & 43.6 \\

				 \bottomrule
			\end{tabular}
		}	
	\end{center}
		\caption{
  {Additional techniques in Argoverse 2 dataset.} 
  $^\ast$: having the same setting with FSDv1, constituting a strictly fair comparison.
  \dag: using a larger Sparse-UNet as the backbone.
  \ddag: using the fading strategy~\cite{pointaugmenting}.}
  \vspace{-3mm}
	\label{tab:argo_roadmap}
\end{table}

\subsection{Runtime Evaluation}
\label{sec:runtime}
Finally, we provide a detailed runtime evaluation to demonstrate the efficiency of FSDv2.
In Table~\ref{tab:runtime}, we list the latency breakdowns of FSDv1 and FSDv2.
The segmentation part encompasses the Sparse-UNet backbone, point-wise classification, and center voting.
As FSDv1 and FSDv2 share the same segmentation structure, they exhibit similar latency in this component.
In FSDv1, the subsequent structures consist of the clustering and SIR module. In contrast, the counterparts in FSDv2 are the virtual voxelization and virtual voxel mixer.
The virtual voxelization is much simpler than clustering, leading to lower latency.
However, it is worth noting that the virtual voxel mixer in FSDv2 is slightly slower than the SIR module in FSDv1. This can be attributed to the fact that in FSDv2, both foreground virtual voxels and all real voxels are taken as inputs to the mixer.
In summary, FSDv2 is on par with FSDv1 in terms of efficiency and achieves significant performance gain, striking a balance between accuracy and computational efficiency.

\begin{table}[H]
\setlength{\tabcolsep}{3pt}
	\begin{center}
		\resizebox{0.99\columnwidth}{!}{
			\begin{tabular}{l|c|cccccc}
				\toprule
			
				\multirow{2}{*}{Methods} & \multirow{2}{*}{mAP} & \multirow{2}{*}{Total} & \multirow{2}{*}{Seg.} & \multirow{2}{*}{\shortstack[1]{Cluster or \\ Vir. Voxel}} & \multirow{2}{*}{\shortstack[1]{SIR or \\ Mixer}} & \multirow{2}{*}{Head} & \multirow{2}{*}{Others} \\
				&  &  &  &  &\\
				\midrule
				FSDv1 & 34.2 & 82.2 & 45.3 & 14.2 & 6.3 & 6.0 & 10.4\\
    			\textbf{FSDv2} & 36.2 & 79.7 & 45.1 & 5.5 & 12.7 & 5.8 & 10.6 \\
				 \bottomrule
			\end{tabular}
		}	
	\end{center}
		\caption{{Latency breakdowns.} Latency (in milliseconds) is evaluated in a single 3090 RTX GPU with PyTorch implementation. No runtime optimizations (e.g., model quantization, kernel fusion). We report the average latency of 1000 samples with batch size 1 in the Argoverse 2 validation split.}
  \vspace{-3mm}
	\label{tab:runtime}
\end{table}

\section{Conclusion}
In this paper, we present FSDv2, an improved version of the previous fully sparse detector FSDv1.
FSDv2 boosts its predecessor by replacing clustering-based instance-level representation with virtual voxels mechanism, eradicating the strong inductive bias stemming from the handcrafted instance-level presentation.
Thus FSDv2 achieves better general applicability in diverse scenarios, outperforming FSDv1 and recent state-of-the-art methods in three mainstream large-scale datasets.
We have made the complete code publicly available to facilitate the development of fully sparse detectors.


%


\ifCLASSOPTIONcaptionsoff
  \newpage
\fi



%



{
\bibliographystyle{IEEEtran}
\bibliography{references}
}

%





\end{document}